\documentclass[10pt,twocolumn,letterpaper]{article}

\usepackage{cvpr}              

\usepackage{graphicx}
\usepackage{amsmath}
\usepackage{amssymb}
\usepackage{booktabs}

\usepackage[pagebackref,breaklinks,colorlinks]{hyperref}

\usepackage[capitalize]{cleveref}
\crefname{section}{Sec.}{Secs.}
\Crefname{section}{Section}{Sections}
\Crefname{table}{Table}{Tables}
\crefname{table}{Tab.}{Tabs.}

\def\cvprPaperID{*****} 
\def\confName{CVPR}
\def\confYear{2022}

\begin{document}

\title{Face-Dubbing++: Lip-Synchronous, Voice Preserving Translation of Videos}

\author{Alexander Waibel\textsuperscript{1,2} \qquad Moritz Behr\textsuperscript{1} \qquad Fevziye Irem Eyiokur\textsuperscript{1} \\ Dogucan Yaman\textsuperscript{1} \qquad Tuan-Nam Nguyen\textsuperscript{1} \qquad Carlos Mullov\textsuperscript{1} \qquad Mehmet Arif Demirtas\textsuperscript{3} \\ Alperen Kantarcı\textsuperscript{3} \qquad Stefan Constantin\textsuperscript{1} \qquad Hazım Kemal Ekenel\textsuperscript{3} \\
\textsuperscript{1}Karlsruhe Institute of Technology, \textsuperscript{2}Carnegie Mellon University, \textsuperscript{3}Istanbul Technical University\\
{\tt\small \{firstname.lastname\}@kit.edu, \{demirtasm18, kantarcia, ekenel\}@itu.edu.tr}
}
\maketitle

\begin{abstract}
   In this paper, we propose a neural end-to-end system for voice preserving, lip-synchronous translation of videos. The system is designed to combine multiple component models and produces a video of the original speaker speaking in the target language that is lip-synchronous with the target speech, yet maintains emphases in speech, voice characteristics, face video of the original speaker. The pipeline starts with automatic speech recognition including emphasis detection, followed by a translation model. The translated text is then synthesized by a Text-to-Speech model that recreates the original emphases mapped from the original sentence. The resulting synthetic voice is then mapped back to the original speakers' voice using a voice conversion model. Finally, to synchronize the lips of the speaker with the translated audio, a conditional generative adversarial network-based model generates frames of adapted lip movements with respect to the input face image as well as the output of the voice conversion model. In the end, the system combines the generated video with the converted audio to produce the final output. The result is a video of a speaker speaking in another language without actually knowing it. To evaluate our design, we present a user study of the complete system as well as separate evaluations of the single components. Since there is no available dataset to evaluate our whole system, we collect a test set and evaluate our system on this test set. The results indicate that our system is able to generate convincing videos of the original speaker speaking the target language while preserving the original speaker's characteristics. The collected dataset will be shared.
\end{abstract}

\section{Introduction}
\label{sec:intro}

Speech-to-Speech translation systems have matured in recent years from early prototypes over mobile hand-held translators to fully integrated and operational simultaneous interpreting systems that have been deployed in lecture and video conferencing applications~\cite{waibel1991janus,lavie1997janus,fugen2007simultaneous,kolss2008simultaneous,eck2010jibbigo,waibel2012simultaneous,muller2016lecture}.
They have proven quite effective in practical deployments and commercial operations using different delivery mechanisms and modalities appropriate to their use case.
In mobile consecutive translation of dialogues (travelers, healthcare providers, humanitarian missions, etc.) individual sentences are translated and the output is commonly synthesized in a target language.
Simultaneous interpretation of lectures, movies and video conferences by contrast are best delivered by subtitling~\cite{muller2016lecture,waibel2016hybrid}, as they can be generated simultaneously~\cite{niehues2018low,nguyen2020super,pham-etal-2021-multilingual,nguyen-etal-2021-kits,anastasopoulos2022findings} and do not create distractions during a speech or monologue.
Still, when movies or off-line video recordings are to be produced, subtitling is sometimes tiresome and a distraction of its own.
Movies, therefore, are sometimes also "dubbed" as an alternate form of delivery, where voice talents act out translated sentences in a target language to replace the original voice.
So far such dubbing has been produced only for movies after the fact but it is costly, requires considerable human effort, and the result is frequently not convincing when the original video and the target voice and language don't properly align.
One proposed solution to improve on these problems is to apply \textit{isometric} human or machine translation \cite{lakew2021isometricmt,anastasopoulos2022findings}, where speech translation is performed on an original video source in a manner that optimizes a temporal match between the translator's generated output text and the original video.
With isometric translation a better dubbing could thus be achieved, but the dubbed speech from a voice talent (or synthetic voice) in the output language still does not match well with the lip movement and the voice of the original speaker in the original video.

In this paper, we propose a different approach:
Rather than inserting translated speech into the original video, we modify the original video in such a way that the resulting video shows lip movements corresponding to the \textit{translated} speech, and the translated synthetic speech in the target language is also generated in a way that is preserving the original speaker's voice characteristics.
The result is a more convincing video experience in the \textit{target} language as \textit{lips} and \textit{voice} match \textit{speech} and \textit{speaker}.
While this idea had already been proposed before in early work on a face translator \cite{ritter1999face}, the integration was not smooth and unconvincing, and only a different synthetic voice could be generated.
Important component technologies, however, have advanced considerably so that a complete more convincing integrated face translator architecture can now be realized that produces a more realistic and convincing experience.
These include large vocabulary low latency simultaneous speech recognition and translation systems \cite{muller2016lecture}, voice conversion \cite{wang2021vqmivc} and various forms of video manipulation \cite{Wav2Lip} and lip syncing \cite{syncnet} methods.  
In the following, we propose an architecture that builds on these advances for an end-to-end speech translation system with voice conversion and lip synchronization that is able to take videos of English-speaking subjects in real-time and generates videos of these speakers with translated German audio and adapted lip movements while also preserving the original speaker's voice characteristics, prosodic cues and emphases. 

In the following, we do not only develop and explore the different component models, e.g., Automatic Speech Recognition (ASR), Machine Translation, Text-to-Speech Generation (TTS), Voice Conversion, and Lip Generation, but also investigate how to employ these models together to provide an effective and accurate end-to-end video experience to translate speech to a target language and generate the synchronized lips with respect to the translated audio data. Several challenging issues must be addressed.
First, we need to provide robust ASR processing to prevent any loss in the original content but preserving details in prosody for better naturalness downstream.
Second, we should have an effective translation system to translate the transcribed text from input language to the target language without error and missing content while moving emphasis information recognized during ASR to the appropriate parts of the translated speech.
Another important dimension is the ability to generate natural, synthetic audio from the translated speech in another language, but preserving the original speaker's voice.
For this, a TTS system is designed to work accurately from translated text data while providing means for fine-grained prosody control. These means of prosody control are then used to add emphases to the generated speech that match the emphases in the original speech.
After speech is generated, we must use voice conversion to adapt the TTS output back to the input speaker's voice, since a TTS is trained on a single or multiple but fixed speakers and thus cannot generate speech with arbitrary voices.
During this adaptation, voice conversion must not cause any degradation in speech quality.
Finally, we need to generate the lips with respect to the translated, voice-converted speech.
During this, speaker identity must be preserved which means the image generation model should not cause any degradation on the face and lips while generating output.
Last but not the least, we need to run all these models sequentially and in a pipelined fashion using the outputs of previous models as an input to the next with minimum delay and without degradation in performance of each model in order to provide a robust end-to-end system.

Our multimodal system includes two pipelines:
a video pipeline for face detection and lip synchronization, and an audio pipeline for speech recognition, translation, speech synthesis, and voice conversion.
The desired output of the audio pipeline is audio of the original speaker uttering a translation of the speech in the input video with properly aligned emphases if any are present in the original audio. This is achieved by pipelining multiple models.
First, our ASR model with emphasis detection creates a transcript of the original speech with additional emphasis information.
Then, the English transcript is translated to German by our translation model while any emphasis information is moved to the corresponding parts of the German translation.
Now, our TTS model synthesizes German speech with appropriate emphases for the given translation and the voice conversion model adapts the synthesized speech to the voice characteristics of the original speaker. Meanwhile, the video pipeline gets the input video frames to detect the speaker's face in them. Finally, the lip generation module employs the generated speech and detected faces to synthesize new frames of the speaker's face with lips that are synchronized to the generated speech.
To evaluate our system, we conducted comprehensive experiments to evaluate the performance of each module as well as the entire system. 

To asses the effectiveness of the resulting system, we collected a test set that contains 262 videos belonging to 25 different speakers.
We carried out a user study to study different aspects of output quality including intelligibility and naturalness of speech, synchronicity of lips and audio, and the credibility of the generated faces in the video.\\

In this paper, we propose a novel architecture that combines recent advances and new techniques in an end-to-end system that achieves the dream of language transparent communication, i.e. creating a video communication experience (in audio and video) between people speaking different languages that removes the language barrier. More specifically, :

\begin{itemize}
    \item We propose an integrated neural end-to-end system to perform automatic video translation that creates the illusion of a speaker speaking another language. Given the video of a speaker, a translated and high-quality lip-synced version of the video is generated that preserves emphases, prosody, the face and voice characteristics of the original speaker.
    \item A real-time, low latency end-to-end speech translation system capable of translating speech from many languages to text in many others (subtitling) is extended for synthetic speech output.
    \item We present a variation to the FastSpeech 2 TTS model that generates synthetic speech but also permits fine-grained prosodic control for the synthesized speech so as to retain emphasis and prosody of the original speech.
    \item A voice conversion module is developed and deployed that maps the synthetic speech back to the voice of the original speaker, even though no data from that speaker is available in the target language.
    \item We develop a real-world dataset to evaluate the components and the overall system. It contains 262 videos of 25 different speakers. The dataset will be shared upon publication.
\end{itemize}

\section{Related Work}

\subsection{ASR and MT}
Attention-based models based on sequence-to-sequence (S2S) \cite{DBLP:journals/corr/abs-2010-03449,nguyen2020super,ansari-etal-2020-findings,pham2022adaptive} are currently one of the top-performing approaches to end-to-end ASR and MT.
A significant amount of study has already been spent to improving the performance of S2S models.
Attention-based S2S models, which use a neural network architecture to approximate the direct mapping from the input to the textual transcript, have become a very efficient approach for building high performance speech recognition systems or machine translation systems, with a very low real-time factor and a significantly lower word error rate in batch processing on GPUs.
The S2S technique has the benefit of simplifying the training of a full end-to-end system, hence hiding the knowledge of complicated components, as in statistical ASR or MT systems.
The detail of our S2S ASR and MT system is describe in Section \ref{ASR}.

\subsection{TTS}
Generating text from speech is a mature research field. Recent developments show that here too that deep learning approaches are effective to generate superior quality speech when compared with traditional approaches. Tacotron \cite{wang2017tacotron}, Fastspeech \cite{ren2020fastspeech} and others now outperform traditional approaches as general-purpose TTS systems and are trainable on raw speech data with transcripts. Since TTS is a hard task due to its inherent one-to-many mapping problem, most modern TTS models are using Mel spectrograms as an intermediate target. The one-to-many mapping problem in TTS refers to the fact that for a given text, there are a large number of possible audio sequences that can be considered fitting TTS outputs as there are many valid variations in voice, prosody, and background noise. To turn the Mel spectrogram outputs into audio waveforms, a vocoder model is subsequently used. Tacotron 2 \cite{shen2018tacotron2}, and many subsequently published variations of it, remain widely used TTS models. Its architecture, an encoder-decoder sequence-to-sequence model based on recurrent units, has the drawback that it is auto-regressive, which makes it hard to parallelize the inference process. To accelerate inference, multiple alternative TTS models with non-autoregressive architectures have been proposed. One of these is FastSpeech 2 \cite{ren2020fastspeech} which does not employ recurrent units, while providing slightly better audio quality than other state-of-the-art models like Tacotron 2 \cite{shen2018tacotron2}, as evaluated by a survey in the original paper. This also makes it easier to run it incrementally so as to minimize resulting latency. We have modified the FastSpeech 2 architecture to provide fine-grained prosodic control to be able to use information about emphases from the original speaker's voice during speech synthesis.

\subsection{Voice Conversion}
The purpose of the Voice Conversion module is to make the resulting voice of the speaker in the target language sound like the original source speaker's voice. Classical Gaussian Mixture Model-based strategies had been proposed and performed well, but modern Artificial Neural Network-based techniques outperform them. GAN, VAE, and seq2seq architectures have been utilized to overcome voice conversion challenges. Voice Conversion systems can be configured in a variety of ways, including one-to-one, one-to-many, many-to-many, any-to-any, and so on. The most challenging voice conversion scenario is given by any-to-any Voice Conversion systems, to convert any source voice to any target speaker, even one not seen in the training data. This has been attempted by several previous architectures such as VQMIVC \cite{wang2021vqmivc}, AutoVC \cite{qian2019autovc}, Adain \cite{chou2019one}, FragmentVC \cite{lin2021fragmentvc}. According to r benchmark comparisons, the VQMIVC is one of the best any-to-any voice conversion systems. We describe the VQMIVC in detail in Section \ref{Voice Conversion}. Finally, we train VQMIVC for converting our TTS output back to the voice of our input speaker.

\subsection{LipSync Video Generation} 
Perhaps the earliest attempt at lip generation from text in a foreign language was presented by Ritter et al.\cite{ritter1999face}. In their work video was synthesized for speakers speaking another language with lip movement synchronous to the speech of the other language. However, rendering smooth lip motion and integration in a face for convincing natural videos was not yet possible. Articulation still appeared jumpy and unnatural and voices were synthetic and differed from the input speaker. Improvements in lip and face synthesis were necessary to create more natural synthetic videos reflecting the original speaker.  

Neural techniques have re-energized new efforts and enabled considerable advances to synthesize video based on arbitrary voice track or text. Many initial efforts were mostly focused on speaker-dependent approaches \cite{ritter1999face,suwajanakorn_synthesizing_2017,kumar_obamanet_2017,fried_text-based_2019,thies_neural_2020}, but more recent efforts \cite{chen2018lip,jamaludin_you_2019} present fully speaker and language independent approaches. \cite{wiles2018x2face} proposed X2Face, which can regenerate the reference video using a variety of modalities, including an input clip or another video to be used as the pose reference. \cite{chen2019hierarchical} separated the audio embedding network from the video generation network to reduce error accumulation. They proposed using attention mechanisms in video generation to achieve higher visual quality than previous methods. \cite{zhou_talking_2019} used language-specific adversarial classifiers to disentangle audio-visual embedding to increase lip-sync quality. \cite{k_r_towards_2019} built upon the architecture in \cite{jamaludin_you_2019} by implementing a language-independent lip synchronization discriminator. \cite{vougioukas_realistic_2020} employed a noise generator in addition to audio and identity encoders in order to capture minor changes in facial expressions that are not captured by the audio. \cite{Wav2Lip} proposed the use of a pretrained classifier as the expert discriminator in \cite{k_r_towards_2019} that provides supervision on the lip-sync accuracy. In our work, we inspire from~\cite{Wav2Lip} to design a lip generation model due to its outperforming lip-sync performance. 

Recent studies focused on not only the lip synchronization but also providing variance in the head pose and head movements of the subject. \cite{zhang_facial_2021} proposed a new GAN-based model that captures implicit attributes related to head pose in addition to lip-sync information. \cite{zhou_pose-controllable_2021} introduced an additional pose source as input to add realistic movement to lip-synced talking heads. Another research direction is to utilize a single reference image instead of a short clip to synthesize videos~\cite{zhou_makelttalk_2020,zhang2021flow,wang_audio2head_2021}. Also, 3D model-based approach~\cite{lahiri2021lipsync3d} and NeRF-based methods~\cite{guo2021ad, yao_dfa-nerf_2022, liu_semantic-aware_2022} are presented to allow for head or body rotation.

 
 \begin{figure*}
    \centering
    \includegraphics[scale=0.30]{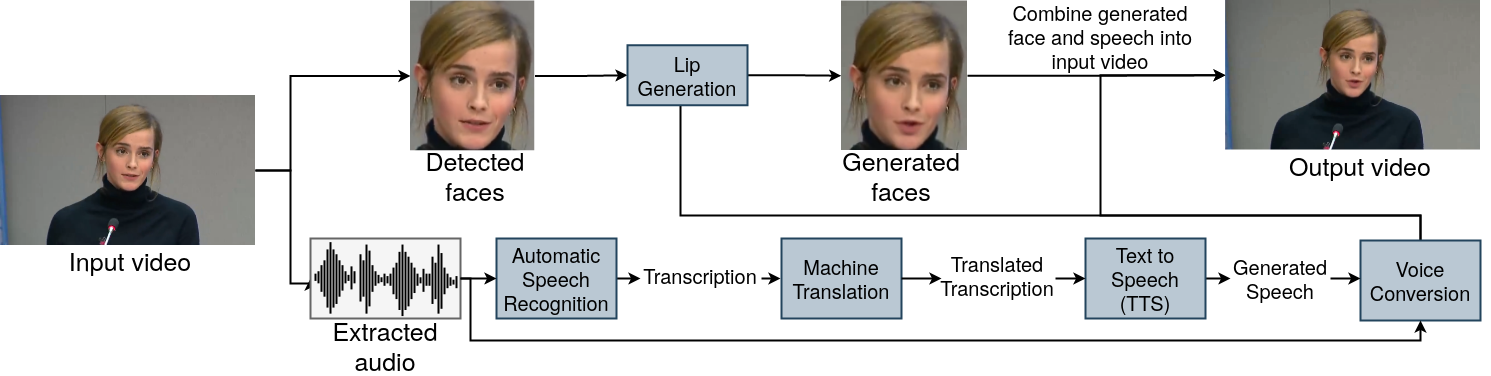}
    \caption{Pipeline of the proposed end-to-end speech-translated lipsync-video generation system. Our system first obtains the audio data from the video input and then extracts Mel spectrogram representation of the audio data. Afterwards, automatic speech translation system provides transcription of the audio to the machine translation system. Later, we acquire the translated text data and send it to the text-to-speech generation system to synthesize the output audio. In order to generate the speech with the same voice of the speaker in the input data, we utilize a voice conversion model and make the synthesized output speech the same with the speaker's voice. Meanwhile, face detector captures faces from each frame and then we provide these faces to the lip generation model in conjunction with the synthesized speech. In the end, we create the video with the synthesized frames and combine with the generated speech to achieve the same video with the input video that has translated speech and synchronized lips.}
    \label{fig:whole_system}
\end{figure*}

\section{System Components}

Our proposed system contains five different modules which are ASR, machine translation, TTS, voice conversion, and lip generation. The system takes an input video and then extracts the audio and video frames. While the ASR model uses extracted audio to transcribe it and detect emphases in it, the translation model receives the transcribed text and detected emphases to translate it to the target language and move the emphases to the appropriate words in the translation. Afterwards, the TTS model generates new audio with appropriate emphases using the translated text emphasis information and sends the output to the voice conversion model. Later, voice conversion adapts the TTS output to the speaker's voice and provides the output to the lip generation model. Meanwhile, the face detection model runs on the video frames to extract faces. In the end, the lip generation system obtains consecutive face images and generated speech, which is the output of the voice conversion system, to synthesize the output face that should have the synchronized lips. An high-level overview of the pipeline is illustrated in Figure~\ref{fig:whole_system}.

\subsection{ASR} \label{ASR}
First, we trained a sequence-to-sequence ASR model to transcribe audio of English (or other language) speech. At our laboratories, three architectures are under investigation: A long short-term memory (LSTM) based model, a Transformer, and a Conformer LSTM-based model. LSTM-based \cite{nguyen2020improving} models include 6 bidirectional layers for the encoder and 2 unidirectional layers for the decoder, with 1536 units in each. They have delivered superior recognition performance on the Switchboard conversational speech benchmark task \cite{nguyen2020super}. 
The Transformer-based model proposed in \cite{pham2019very} feature 24 encoder layers and 8 decoder layers. The Conformer-based model \cite{gulati2020conformer} consists of 16 encoder layers and 6 decoder layers. The size of each layer in both the Transformer-based and the Conformer-based models is 512, while the size of the hidden state in the feed-forward sub-layer is 2048.  As explained in \cite{nguyen2020improving}, the speech data augmentation approach was employed to reduce over-fitting. Also recent work on factorizing multilingual models delivered considerable improvements in view of broad multilingual expansion\cite{pham21_interspeech}. 
In the present implementation we used Stochastic Layers with a dropout rate of 0.5 on both Transformer-based and Conformer-based models to successfully train a deep network \cite{pham2019very}. To classify a word as emphasized, we add a binary classifier layer to the end of the network. The ensemble of LSTM-based and Conformer-based sequence-to-sequence model provided the best results.

\subsection{Translation}
We translate from English to German (and, indeed, to many other languages) using a neural sequence-to-sequence model.
More specifically, we employ a Transformer~\cite{10.5555/3295222.3295349} model with the \emph{base} configuration as described by~\cite{10.5555/3295222.3295349}, implemented in the NMTGMinor framework~\cite{pham20_interspeech}.
We train the model on 1.8 million sentences of Europarl data~\cite{koehn2005europarl} and finally finetune on 150,000 sentences of TED data~\cite{cettolo2012wit3} for better adaptation towards spoken language.

For emphasis translation, we extract a source-to-target word alignment.
For each emphasized input token, we then determine the matching output token and put emphasis on this corresponding output token.
The word alignment we obtain by averaging the normalized attention scores from each head of the final encoder-decoder multihead-attention layer:

\begin{center}
\begin{equation}
\alpha_{ji} = \frac{1}{h}\sum_{k=0}^{h}{\alpha^{k}_{ji}}
\end{equation}

\begin{equation}
\alpha^{k} = \text{softmax}\left(\frac{(QW^Q_k)(KW^K_k)^T}{\sqrt{d}}\right)
\end{equation}
\end{center}

where $ h=8 $ is the number of attention heads, $ d=512 $ is the model size, and $ Q $, $ W $, $ W^K $ and $ W^Q $ follow the description in \cite{10.5555/3295222.3295349}. For each emphasized input token $ s_i $ emphasis is thus put on the output token $ t_j $ with
$ j = argmax_{k = 1..|T|}(\alpha_{ki}) $.

\subsection{TTS} 
We are using a modified FastSpeech 2 \cite{ren2020fastspeech} model for synthesizing Mel spectrograms of speech for a given text. 
We chose FastSpeech 2 over other popular TTS models like Tacotron 2 \cite{shen2018tacotron2} as FastSpeech 2 allows 
for faster inference times due to its non-autoregressive design. Its architecture is based on an encoder-decoder architecture and employs multiple feed-forward Transformer blocks \cite{ren2019fastspeech}
that are made up of stacks of self-attention and TDNN/1-D-convolution layers.

To make non-auto-regressive TTS feasible,
FastSpeech 2 employs variance adaptors which provide information on prosody to ease the one-to-many mapping problem inherent to TTS. 
The three variance adaptors enrich the hidden sequence by adding predicted pitch, duration, and energy information on phoneme-level to the hidden sequence thus helping the decoder by easing the one-to-many
mapping problem of TTS. To further ease the training process of the model and make phoneme-level variance prediction 
possible, the model is given the input text not as a sequence of graphemes but rather as a sequence of phonemes.
Consequently, prior conversion is needed for grapheme inputs. This is done by consulting a pronunciation dictionary
and, for words not present in the dictionary, by employing a grapheme to phoneme model trained using the 
Montreal Forced Aligner \cite{mcauliffe2017montreal}.

Originally, the predictions of the variance adaptors can only be controlled by parameters for the entire utterance which would not allow for fine-grained prosody control. As we aim to add emphases to the synthesized speech that match the emphases in the original speech, we then add prosody controls at the word-level to the text input by way of Speech Synthesis Markup Language \cite{taylor1997ssml} (SSML) tags.  Using SSML, we can now add emphasis tags to words in the translation that correspond to words in the original transcript that were emphasized by the speaker. In our system, this happens automatically as the ASR model adds emphasis tags to text sections where emphases were detected. The prosody predictions of the variance adaptors are then modified for the phonemes of that word to create an emphasis in the TTS output.  The model varies duration and energy of the respective phonemes as well as increasing or decreasing pitch depending on the originally predicted pitch for the word. Finally, we use the HiFi-GAN vocoder \cite{hifigan} to generate audio wave-forms from the Mel spectrograms generated by the TTS model.

\subsection{Voice Conversion} \label{Voice Conversion}
Following TTS in a standardized voice in the target language with the prosody projected from the source speech, we aim to revert the generated speech to the original speaker's voice. To accomplish this, we need to employ voice conversion from the synthetic TTS voice back to the original speaker's voice in our original videos. We use VQMIVC (Vector quantization mutual information voice conversion) as a method for this step. VQMIVC uses a straightforward but effective autoencoder architecture to perform voice conversion in a way that separates the effects of voice from prosody, content and emphasis. The framework consists of four modules: a content encoder that produces a content embedding from speech, a speaker encoder that produces a speaker embedding (D-vector) from speech, a pitch encoder that produces prosody embedding from speech, and a decoder that generates speech from content, prosody, and speaker embeddings, respectively. Phonetics and prosody are represented through content embedding and prosody embedding. The content embedding is discretized by a vector quantization module and used as target for the contrastive predictive coding loss.

A mutual information (MI) loss measures the dependencies between all representations and can be effectively integrated into the training process to achieve speech representation disentanglement. During the conversion stage, the source speech is put into the content encoder and pitch encoder to extract content embedding and prosody embedding. To extract the target speaker embedding, the target speech is sent into the speaker encoder.  Finally, the decoder reconstructs the converted speech using the source speech's content embedding and prosody embedding and the target speech's speaker embedding. 
We adapt the pre-trained VQMIVC voice conversion on both German and English datasets to get better performance on both languages. Our VQMIVC model is fine-tuned with the same hyper-parameters as in the original papers. The evaluation of VQMIVC is presented in \cite{wang2021vqmivc}.

 \begin{figure*}
    \centering
    \includegraphics[scale=0.45]{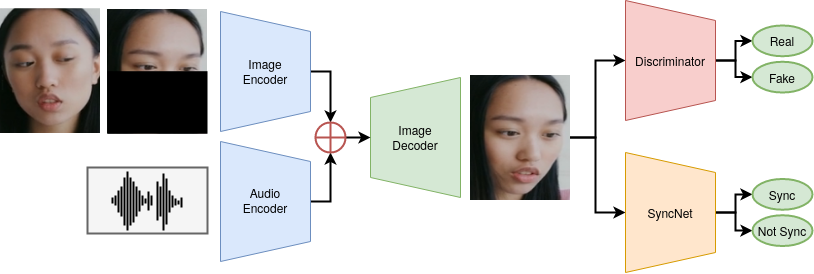}
    \caption{Illustration of the proposed lip generation model. We have two encoders, namely audio encoder and image encoder, and one face decoder to generate face images. After we extract features from the audio and the image, we concatenate them along the depth axis to provide input to the face decoder. Besides, we have a discriminator network to evaluate the quality of the generated face images and decide whether they are real or fake. Finally, we have a pretrained synchronization network that is classifying the generated face image to determine whether it is synchronized with the input audio. Please note that we train the whole video pipeline as end-to-end.}
    \label{fig:lip_generation}
\end{figure*}
 
\subsection{Lip Generation}

We address the lip generation task as a conditional generative adversarial network-based~\cite{vanillaGAN,cGAN} image generation, since our goal is to generate lips with respect to the audio input and face input in order to make the generated lips synchronized with the audio. We design our GAN model with inspiration from \cite{Wav2Lip}.  First of all, we propose an audio-guided face generator $ G $ to synthesize a face image that is synchronized with the audio. For this, we first obtain Mel spectrogram representation of the audio data. Afterwards, we provide an audio sequence as a sequence of Mel spectrogram to our audio encoder as an input. The audio encoder is responsible for embedding the audio input in order to extract the embedded feature representation. Meanwhile, we utilize an image encoder to encode the input image. Our input image has six channels, namely the depth-wise concatenation of two separate images. While the first three channels contain a face of the corresponding subject from another time sequence or from another video of the same subject, namely reference image $ x_r $, the second image is the masked version of the ground truth face, $ x_m $. The task is to generate the masked area of $ x_m $ with respect to the audio sequence. We basically mask the half-bottom part of the face image. Since preserving the identity and details of the face to improve the realism of the final image are crucial, the reference image $ x_r $ is necessary to inject these details to the $ G $ while the final face image is being generated. Otherwise, it would be challenging for our generator to preserve the identity and the details in the bottom part of the image. After we acquire the audio and face feature representations from audio and image encoders, we concatenate them along the depth. We further feed the face decoder with this concatenated feature representation. 

We further utilize residual connections between the reciprocal layers of the image encoder and image decoder networks in our generator $ G $. These connections allow us to transmit the output of encoder's layers to the decoder's layers in order to transfer the crucial details and identity of the input face images. We utilize ReLU activation function in our generator with instance normalization layers.

For the discriminator, we employ a binary classifier with a cross-entropy loss to distinguish real and fake images. This discriminator is responsible for the quality and realism of the generated image. In the discriminator, we benefit from spectral normalization to provide more stable training by normalizing the gradients. Besides, we employ Leaky ReLU and Instance normalization in discriminator. In addition to this, we must also control whether the prior condition is provided in the generated image as it is proposed in~\cite{Wav2Lip}. Therefore, we utilize a pretrained synchronization model~\cite{syncnet, Wav2Lip} to evaluate the coherence between the conditional input audio and the output face image. This synchronization network is also a binary classifier that classifies the image to produce output whether synchronization is provided or not. The whole lip generation model is illustrated in Figure~\ref{fig:lip_generation}. 

To train our system, we employ a large-scale Oxford-BBC Lip Reading Sentence 2 dataset (LRS2)~\cite{bbc1,bbc2,bbc3}. We feed our image generator with a set of five consecutive frames. We further send the audio data to audio encoder after we obtain Mel spectrogram representation of the corresponding audio sequence. During the experiments, we follow the proposed data splits to train, validate, and test our model. In order to calculate synchronization loss, we directly use the pretrained lip synchronization model~\cite{Wav2Lip}, and we do not update this model during the training. Our overall loss function is as follows:

\begin{equation}
	L = L_{cGAN} + \alpha * L_{img}  + \beta * L_{sync}
\end{equation}

where $ L_{cGAN} $ is a conditional adversarial loss, $ L_{img} $ is an image reconstruction loss, $ ||y - y'|| $, that calculates the L1 distance between target face image and the generated face image in the pixel space. $ L_{sync} $ is a synchronization loss that provides feedback to the generator whether the synchronization between the lip and the audio input is able to be provided in the generated face image. $ \alpha $ and $ \beta $ are coefficients that alter the effect of image reconstruction loss and synchronization loss on the total loss. According to the experimental results, we find the best $ \alpha $ and $ \beta $ coefficients as $ 1 $ and $ 0.05 $.

\section{System Integration}

In order to combine the multitude of models for ASR, translation, TTS, voice conversion, and lip generation into a single system, we chose a cascade architecture. A diagram of the high-level architecture of our system is shown in Figure \ref{fig:whole_system}. Initially, the audio of the given video is extracted and converted to the expected waveform format of the ASR module which then creates an English transcription of the input speech with additional information regarding detected emphases. Then, the translation module produces a German translation of that transcript, including SSML tags for emphases at the parts of the text that correspond to words in the original English transcript that were marked as emphasized by the ASR module.
Subsequently, the TTS module is given this translated text and generates German speech with emphases according to the SSML tags. The resulting Mel spectrogram is turned into a waveform file by the HiFi-GAN vocoder. Afterwards, the final audio is created by the voice conversion module which gets the waveform of German speech that the vocoder produced as input and uses the original English audio of the input video as target speaker. Meanwhile, the video pipeline starts by detecting faces in the input video to provide consecutive face images to the lip generation model. Besides, the lip generation module is given the speech produced by the voice conversion model to generate face images with the modified lips. In the end of the system, the generated faces and generated speech are combined to create the final output video. This whole pipeline allows us to acquire a video with translated speech of the original speaker in the target language and the adapted lips by only providing an arbitrary video.

\section{Experimental Results}

\subsection{Dataset} \label{dataset}

We used various datasets to train and evaluate our models. Besides, we collected a test set to measure the performance of the entire system. 

\textbf{ASR and MT training dataset} 
For training and evaluation of our ASR models, we used Mozilla Common Voice v6.1 \cite{ardila2019common}, Europarl \cite{koehn2005europarl}, How2 \cite{sanabria2018how2}, Librispeech \cite{panayotov2015librispeech}, MuST-C v1 \cite{mustc19}, MuST-C v2 \cite{cattoni2021must} and Tedlium v3 \cite{hernandez2018ted} datasets.  We also collected the text parallel training data provided by WMT 2019, 2020, 2021 for training MT consisting of a total of 69.8 million sentences as shown on the right side of  Table \ref{table:asr-english-data}.

\begin{table*}[h]
\centering
\caption{\label{table:asr-english-data} Summary of the English datasets used for speech recognition (left) and machine translation (right)}
\begin{tabular}{l|r|r} \toprule
 Corpus & Utterances & Speech data [h]\\ \toprule
 \multicolumn{3}{l}{\textbf{A: Training Data}} \\
 \toprule
Mozilla Common Voice  & 1225k & 1667 \\ 
Europarl & 33k & 85\\ 
How2 & 217k & 356\\ 
Librispeech & 281k & 963\\ 
MuST-C v1 & 230k & 407\\ 
MuST-C v2 & 251k & 482\\ 
Tedlium & 268k & 482\\ 
\toprule
 \multicolumn{3}{l}{\textbf{B: Test Data}} \\
 \toprule
Tedlium & 1155 & 2.6\\
Librispeech & 2620 & 5.4\\
\hline
\end{tabular}\quad\quad\quad\quad
  \begin{tabular}{lr} \toprule 
    Dataset & Sentences  \\
 \toprule
TED Talks (TED) & 220K  \\
Europarl (EPPS)  & 2.2MK  \\
CommonCrawl & 2.1M  \\
Rapid & 1.21M  \\
ParaCrawl & 25.1M  \\
OpenSubtitles & 12.6M  \\
WikiTitle & 423K  \\
Back-translated News & 26M  \\
\hline 
\end{tabular} 
\end{table*}

\textbf{CSS10 German dataset~\cite{park2019css10}.} CSS10 is a collection of single speaker speech datasets that contain ten different languages. It includes short audio clips and their aligned text data. Since we aimed to generate the audio in German, we utilized the CSS10 German dataset to train our TTS model as it provides 17 hours of high-quality single speaker audio data which is enough to train a single speaker TTS model.

\textbf{LRS2.} We employed the Oxford-BBC Lip Reading Sentences 2 (LRS2) dataset to train our lip generation model and also evaluate its performance. We followed the presented train, validation, and test setups to train the model as well as evaluate the performance. The training set contains 45839 utterances, while validation and test sets include 1082 and 1243 utterances respectively. 

\textbf{Our dataset.} Since there is no suitable dataset to test our end-to-end video translation system in the literature, we collected various videos from the internet to create a test set. Our test set contains 262 different video clips belonging to 25 different speakers. The duration of the test clips is about ten seconds. Please note that all speakers speak English since we evaluated our system for the combination of English input and German output.

\subsection{Evaluation metrics}

\textbf{WER and BLEU}
The word error rate (WER) is a common metric for measuring speech recognition performance. The Levenshtein distance at the word level is used to calculate the WER. The WER of Librispeech test set represents the ASR's performance on read speech, while the WER of Tedlium test set represents the ASR's performance on spontaneous speech. The BLEU, or Bilingual Evaluation Understudy, is a score that compares a candidate translation of text against one or more reference translations \cite{papineni-etal-2002-bleu}.

\textbf{LSE-D and LSE-C~\cite{Wav2Lip}.} Since the FID, SSIM, and PSNR are not able to evaluate the synchronization of the lips and the synchronization is a crucial key-point in the lip generation task in addition to the quality of the generated face images, using Lip-Sync Error-Distance (LSE-D) and Lip-Sync Error-Confidence (LSE-C) provide more reliable representation about the synchronization. Therefore, as it is proposed in~\cite{Wav2Lip}, we utilized LSE-D and LSE-C metrics to evaluate the synchronization performance of our lip generation model. 

\textbf{FID~\cite{FID_score}.} In order to evaluate the quality of the generated face images, we employed FID score~\cite{FID_score} by providing the manipulated face images. Thus, FID basically calculates the distance between real samples and generated samples in the feature space. For this, Inceptionv3 image classification model, that was trained on ImageNet dataset, is utilized to extract features. In this metric, lower score indicates better quality for the generated images.

\textbf{User study.} For the evaluation of TTS model as well as the whole system there are no widely accepted computable quality metrics. So in order to evaluate the TTS model and the whole system, we conducted user studies and asked participants to evaluate the performance in several different aspects.


\subsection{ASR and Translation}

Our ASR and translation models are evaluated by employing computable metrics on standard datasets. For ASR, our ensemble of LSTM-based and Comformer-based sequence-to-sequence model achieves WERs of respectively 2.4 and 3.9 on  the Libri and Tedlium test sets. In Table \ref{res:small}, we present the results of Conformer-based, Transformer-based, LSTM-based, and ensemble-based approaches. According to the table, ensemble-based method achieves the best results on Libri test, while it reaches the same performance with LSTM-based approach and surpass the Conformer-based and Transformer-based methods on TED-LIUM test set. Therefore, we decide to use our ensemble-based approach in the final proposed system. Besides, our translation model attains a translation score of 29.7 BLEU on the IWSLT \textit{tst2010} test set.

\begin{table}[htb]
\caption{\label{res:small} WER results on Libri and Tedlium test sets. While we obtain the best result with ensemble-based method on Libri dataset, we get the best results with ensemble-based and LSTM-based methods on Tedium dataset. }
 \begin{center}
   \begin{tabular}{lcc} \hline 
    Data & Libri & Tedlium \\
\toprule
Conformer-based & 3.0 &  4.8 \\
Transformer-based & 3.2 & 4.9 \\
LSTM-based &  \textbf{2.6} &  \textbf{3.9} \\
\toprule
Ensemble &  \textbf{2.4} & 3.9 \\
\bottomrule
\end{tabular}   
\end{center}
\end{table}

\subsection{TTS}
We trained our TTS model on the CSS10 German dataset~\cite{park2019css10} which is a single-speaker dataset consisting of nearly 17 hours of German speech and on the LJSpeech \cite{ljspeech17} dataset, an English single-speaker dataset consisting of approximately 24 hours of speech.
Montreal Forced Aligner was used to transform the grapheme inputs of the datasets to phoneme sequences and generate the
text-audio alignments needed for training the variance adaptors. Training was done on a server with an Intel 4124 CPU, 32 gigabytes of memory,
and a single NVIDIA RTX Titan GPU and took approximately 72 hours.
A pretrained universal HiFi-GAN model was used as vocoder, no finetuning was necessary.

The evaluation of the TTS system was done in two user studies. A first study was conducted to compare the performance of our modified FastSpeech 2 architecture on the LJSpeech dataset with the widely used Tacotron 2 architecture to get a baseline. A second user study was done on our model which was trained on the German CSS10 dataset in order to evaluate it's performance when applying fine-grained prosody control.

For comparison with Tacotron 2, we synthesized ten texts from the test set of the LJSpeech dataset with both Tacotron 2 and FastSpeech 2. For ground truth comparison we used the respective audio samples. A group of eight participants was then asked to rate the quality of the audio samples on a scale from 1 to 5. After that, mean opinion scores (MOS) and confidence intervals were calculated. Table \ref{table:eval_eng_tts} shows the MOS and confidence intervals results from this survey. As the results show, our modified FastSpeech 2 model performs as well as Tacotron 2. This confirms the results of the FastSpeech 2 evaluation in \cite{ren2020fastspeech} and suggests that our modifications to FastSpeech 2 did not decrease the quality of the synthesized speech.

\begin{table}[t]
\centering
\caption{\label{table:eval_eng_tts} MOS and 95\% confidence intervals for ground truth samples and TTS syntheses by Tacotron 2 and modified FastSpeech 2.}
\begin{tabular}{c|c}\hline
                       & MOS           \\ \hline
Ground Truth           & 4.21 $ \pm $ 0.17 \\ 
Tacotron 2             & 3.86 $ \pm $ 0.21 \\ 
Modified FastSpeech 2 & 3.87 $ \pm $ 0.2 \\
\bottomrule
\end{tabular}
\end{table}

For subjective evaluation of the German TTS system and the fine-grained prosody control capabilites of our model, speech was synthesized for texts randomly drawn from the test set of the CSS10 dataset. For ground truth comparison we further chose random audio samples from the test set. This time we also compared the quality of the generated speech when using default prosody with the quality of generated speech with added emphases. We conducted this additional comparison only on the German model as this is the model we also use in the final system evaluation. To evaluate the capability of the system to add emphases to the synthesized speech, the chosen text samples
were synthesized again, this time with an emphasis added to a random word. To get a more differentiated view on quality differences between unemphasized and emphasized TTS outputs, the group of eight participants asked to rate the audio quality considering two metrics, naturalness and intelligibility on a scale from 1 to 5. For the emphasized TTS outputs perceptibility of emphasis was additionally  rated by the participants. Table \ref{table:eval_tts1} shows the MOS and confidence intervals for ground truth and unemphasized samples.
Table \ref{table:eval_tts2} shows the MOS and confidence intervals for the synthesized samples with added random emphasis. Additionally, changes in naturalness and intelligibility scores when compared with non-emphasized TTS samples are shown.

\begin{table}[t]
\centering
\caption{\label{table:eval_tts1} MOS and 95\% confidence intervals for ground truth and TTS samples.}
\begin{tabular}{c|c|c}\hline
& Naturalness   & \multicolumn{1}{l}{Intelligibility} \\
\hline
Ground Truth  & 4.28 $\pm$ 0.12 & 4.82 $\pm$ 0.08 \\ 
Synthesis & 3.59 $\pm$ 0.28 & 4.69 $\pm$ 0.09 \\
\bottomrule
\end{tabular}
\end{table}

The results show no clear difference between intelligibility scores of synthesized samples and ground truth samples.
However, naturalness is rated worse for synthesized samples, implying a perceptible difference in audio quality or prosody
when comparing ground truth and synthesized samples. But these differences do not seem to decrease intelligibility in any way. Adding emphases to the generated speech seems to slightly decrease naturalness, suggesting that the emphases, while being well perceptible, might not sound entirely natural.

\begin{table}[t]
\centering
\caption{\label{table:eval_tts2} MOS, Comparison, and 95\% confidence intervals regarding naturalness, intelligibility and perceptibility of emphasis for TTS samples with randomly emphasized word.}
\begin{tabular}{c|c|c} \hline
& \begin{tabular}[c]{@{}c@{}}TTS with \\ Emphasis
\end{tabular} 
& \begin{tabular}[c]{@{}c@{}}Change vs. \\ Standard Synthesis
\end{tabular}  \\ \hline
Naturalness & 3.29 $\pm$ 0.16  & - 0.3  \\ 
Intelligibility & 4.71 $\pm$ 0.13 & + 0.02 \\ 
Emphasis & 3.75 $\pm$ 0.28  & - \\
\bottomrule
\end{tabular}
\end{table}

The overall performance of our model, even when emphases are added, is comparable with the MOS results for Tacotron 2 and our modified FastSpeech 2 obtained in the first user study as shown in Table \ref{table:eval_eng_tts}. However, there cannot be any conclusive comparison as these models have been trained for English speech and only a single MOS value was given by the participants.

\begin{table*}
\centering
\caption{\label{table:eval_lip1} Evaluation of Wav2Lip and our model. We test both model on LRS2 test set and our test set. Since we do not have the ground truth outputs for the German language scenario, we could not calculate FID scores. Wav2Lip-GAN results on LRS2 datasets are taken form the corresponding paper~\cite{Wav2Lip}. }
\begin{tabular}{l|c|c|ccc}\hline
Model & Data & Language & LSE-D & LSE-C & FID \\ \hline
Wav2Lip-GAN & LRS2 & English & \textbf{6.46} & \textbf{7.78} & \textbf{4.44} \\
Ours & LRS2 & English & 6.98 & 6.93 & 8.86 \\ \hline
Wav2Lip-GAN & Ours & English & 8.35 & 6.40 & \textbf{19.62} \\ 
Ours & Ours & English & \textbf{8.11} & \textbf{6.52} & 21.15 \\ \hline
Wav2Lip-GAN & Ours & German & 7.93 & \textbf{7.18} & - \\ 
Ours & Ours & German & \textbf{7.90} & 7.17 & - \\
\bottomrule
\end{tabular}
\end{table*}

\begin{figure*}
\small
\begin{center}
\begin{tabular}{cc}
\includegraphics[scale=0.14]{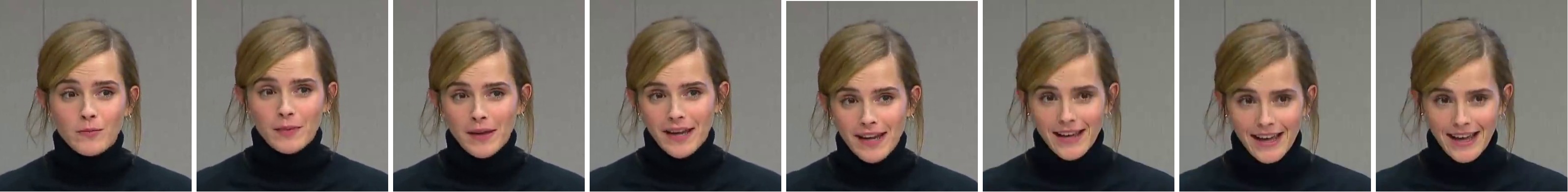}&\\
"Subset of the face images from original video." \\
\includegraphics[scale=0.14]{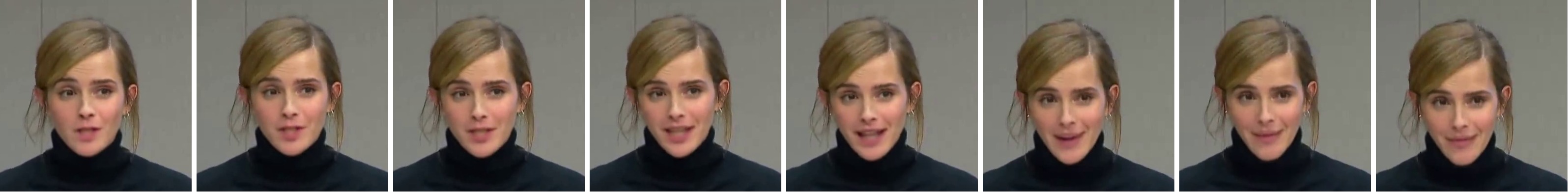}&\\
"Subset of the generated face images based on the translated and generated speech." \\
\end{tabular}
\end{center}
\caption{Sample face images from original video and generated video. In the first row, eight consecutive frames from original video are presented. In the second row, the same eight frames with new lips are presented. The images in the second row are synthesized by using generated German speech which is the translation of the input speech.} 
\label{fig:generated_frames}
\end{figure*}

\begin{figure*}
\small
\begin{center}
\begin{tabular}{ccc}
\includegraphics[width=4.3cm]{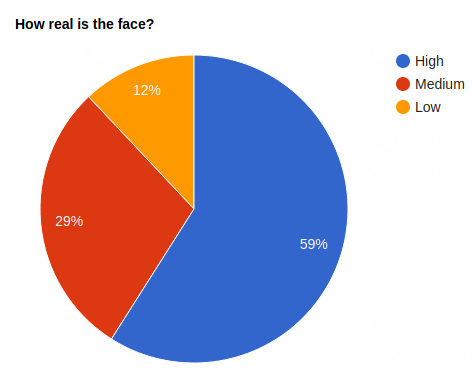}&
\includegraphics[width=4.3cm]{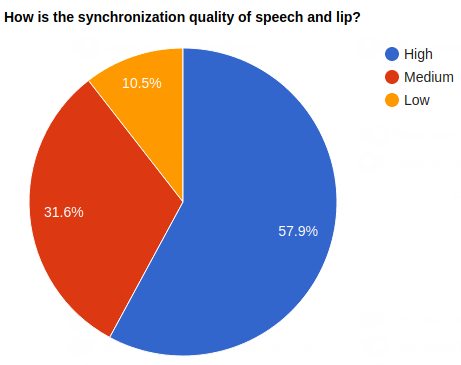}&
\includegraphics[width=5.0cm]{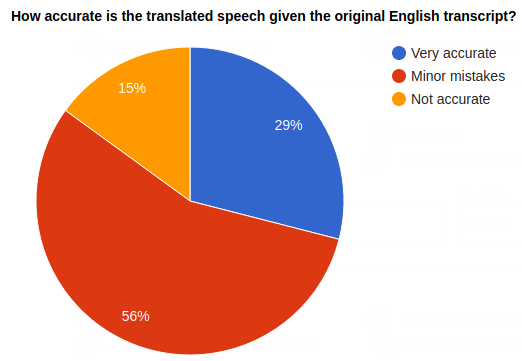}
\end{tabular}
\end{center}
\caption{Subjective test results on our proposed test set. We ask participants to evaluate the generated videos in several different aspects, namely the quality of the generated face images, the synchronization quality of the lips and speech, the accuracy of the translated speech, naturalness of the generated speech, and the intelligibility of the generated speech.}
\label{fig:pie_graph}
\end{figure*}


\subsection{Lip Generation Results}

In order to evaluate the lip generation performance, we follow three different strategies. We first evaluate the quality of the generated images by using FID score~\cite{FID_score}. We further consider the conditional image generation by measuring the synchronization between the generated lip and the audio input. For this, we benefit from recently proposed novel metrics, LSE-D and LSE-C~\cite{Wav2Lip}, which are basically distance and confidence scores for the synchronization performance. Finally, we perform subjective tests to quantify the proposed system's performance. In Table~\ref{table:eval_lip1}, we show LSE-D, LSE-C, and FID scores for LRS2 test set as well as our proposed test set. The LSE-D and LSE-C results on LRS2 dataset show that our model and Wav2Lip~\cite{Wav2Lip} achieve almost the same performance to provide synchronized lips, although Wav2Lip shows slightly better scores. On the other hand, on our dataset, we achieve a slightly better scores in English case and in German case, though the scores are quite similar. This outcome indicates that both models have an effective generalization capacity and they are robust against real-world challenges, since both models show a well performance on unseen dataset. Moreover, FID scores are again very close to each other. However, Wav2Lip achieves better FID scores on LRS2 and our dataset. This FID analysis indicates that the generation quality of our model should be improved. Please note that we could not calculate the FID score for our dataset, since we generate the faces based on the German audio input, therefore, there are no ground truth images.

\subsection{System Evaluation}

In order to evaluate the whole system, we conduct a user study with 25 participants. In this way, we aim to investigate the performance of the system by considering several different aspects: 1) realism of the generated face, 2) naturalness of the generated voice, 3) intelligibility of the speech, 4) synchronization quality of the speech and lip, 5) accuracy of the translated speech given the original English transcript. We ask one question for each aspect. However, please note that only the participants who know the German language answered the questions related to intelligibility and the German translation of the speech. In the user study, we randomly choose 80 videos from the 262 videos of the dataset described in Subsection \ref{dataset} and show the translated and lip-synced results to the participants with the transcribed original speech as well as the five questions. Sample evaluation videos can be found here \footnote{https://videospeechtranslation.github.io}. 

We illustrate results of quality of the generated faces, synchronization accuracy, and the translation accuracy in Figure~\ref{fig:pie_graph}. The results indicate that participants rate the quality of the generated faces as high. Similarly, the majority of the answers state that our system is provides accurate synchronization in the generated videos. For the translation accuracy, although a majority of the answers indicate that there are minor mistakes in the text, only 15\% of the answers find the results inaccurate. Moreover, we demonstrate naturalness and intelligibility results in Table~\ref{table:eval_lip2}. The results demonstrate that our system is successful in providing naturalness and intelligibility in the generated video.

The evaluation showed that the faces and lip-synchronization in the generated videos were believable and the generated speech well intelligible. Sample images are shown in Figure \ref{fig:generated_frames}. However, we observed occasional problems with naturalness of the generated speech and inaccuracies in the translations due to lacking punctuation in the transcripts generated by the ASR model. Moreover, the lip-syncing model showed slight issues with bearded faces and also had some quality problems that must be addressed to improve the quality of the generated faces to make them more natural.

\begin{table}[t]
\centering
\caption{\label{table:eval_lip2} Subjective evaluation on our proposed test set. In the questions, the minimum score is 1 while the maximum score is 5. Scores show the mean value and standard deviation. }
\begin{tabular}{l|c}\hline
Measurement & Score \\ \hline
Naturalness & 3.36 $ \pm $ 0.98 \\ 
Intelligibility & 4.24 $ \pm $ 0.86 \\
\bottomrule
\end{tabular}

\end{table}

\section{Conclusion}

In this work, we proposed an end-to-end system for combined face, lip, audio translation from input video. Given a video of a speaker, our system can generate a convincing output video of that speaker
uttering a translation of the original speech while adapting lip movements to the new audio and preserving voice characteristics. Additionally, emphases are preserved by emphasis detection in the ASR model, and modifications to the used FastSpeech 2 TTS model allow fine-grained prosody control which is used to create corresponding emphases in the synthesized speech. The detailed experimental results of each module and user study for the system evaluation indicated that we achieve accurate modules for each task and acceptable performance in the final system to do the video translation. To address remaining translation issues discovered in our experiments and to improve naturalness of the generated speech preserving pauses, we employ more advanced ASR models with punctuation capabilities and voice activity detection to mark pauses in the transcript. This information improves translation quality and naturalness. Improvements towards more robust voice conversion are also desirable as we still observe occasional robustness issues on long speech inputs. These issues are likely to improve with additional training data that specifically incorporates long sentence speech samples.
Lastly, as the pipeline of our system contains many components, inference times and latency of the ensemble need to be improved. Ongoing work is devoted to improving speed and latency of the components, parallelizing component processing and a better pipelined architecture.


{\small
\bibliographystyle{ieee_fullname}
\bibliography{egbib}
}

\end{document}